\documentclass{article}
\usepackage[utf8]{inputenc}
\usepackage{graphicx} % Required for inserting images
\usepackage[style=authoryear, natbib=true, maxcitenames=2, uniquename=false, uniquelist=false]{biblatex}
\usepackage{subcaption}
\usepackage{booktabs}
\usepackage{placeins} 
\usepackage{hyperref} 
\usepackage{amsmath, xparse}
\usepackage{graphicx}
\usepackage{tabularx}

\title{The Material Contracts Corpus}
\author{Peter Adelson\thanks{Stanford Graduate School of Business and Stanford Law School.} \and Julian Nyarko\thanks{Stanford Law School.}\\
   }

\date{March 31, 2025}

\begin{document}

\maketitle

\begin{abstract}
This paper introduces the Material Contracts Corpus (MCC), a publicly available dataset comprising over one million contracts filed by public companies with the U.S. Securities and Exchange Commission (SEC) between 2000 and 2023. The MCC facilitates empirical research on contract design and legal language, and supports the development of AI-based legal tools. Contracts in the corpus are categorized by agreement type and linked to specific parties using machine learning and natural language processing techniques, including a fine-tuned LLaMA-2 model for contract classification. The MCC further provides metadata such as filing form, document format, and amendment status. We document trends in contractual language, length, and complexity over time, and highlight the dominance of employment and security agreements in SEC filings. This resource is available for bulk download and online access at https://mcc.law.stanford.edu.

\end{abstract}

\section{Introduction}

This paper introduces the Material Contracts Corpus (MCC), a collection of 1,038,766 contracts filed with the Securities and Exchange Commission (SEC) by public companies. This work contributes to ongoing efforts to compile large datasets of contracts to promote empirical studies of contract design, usage, and trends~\citep{adelson2025introducing, chicago, Nyarko2019}. Such datasets have already proven useful for testing theories of private ordering, for example, in international arbitration~\citep{Nyarko2019}. The data set is further intended to assist with the development and evaluation of modern computational methods, such as large language models, for contracting~\citep{guha2023legalbenchcollaborativelybuiltbenchmark, wang2025acordexpertannotatedretrievaldataset, choi2024lawyering, arbel2024generative, koreeda2021contractnli}.

In addition to its scale and temporal coverage, the MCC offers two additional advantages. First, by leveraging open-source artificial intelligence (AI) tools, the MCC provides further details about its contracts, such as the agreement and document type. Contracts can be filtered based on whether they are, for example, employment agreements or leases. Second, the MCC links contracts to specific parties, enabling researchers and practitioners to more easily identify the publicly filed contracts in which a specific legal entity is a party. The MCC is available for both bulk download and online search at https://mcc.law.stanford.edu/.

This paper proceeds as follows: \hyperref[sec:sec2]{Section 2} describes the process of creating the MCC, while \hyperref[sec:sec3]{section 3} outlines certain high-level trends observed within the MCC. A last section concludes briefly.

\section{Creating the Material Contracts Corpus}\label{sec:sec2}

\subsection{Collecting the Contracts}

The SEC requires publicly registered companies to submit their ``material contracts'' to the Electronic Data Gathering, Analysis, and Retrieval (EDGAR) Database—a centralized repository. A ``material contract'' is defined as an agreement `not made in the ordinary course of business that is material to the registrant.'\footnote{17 C.F.R. § 229.601(b)(10)(i).} In practice, these contracts tend to be of significant importance to the filing companies, and may include M\&A agreements, C-suite employment agreements, licensing or joint venture agreements with significant stakes, pension plans etc. The contracts are submitted as exhibits to disclosure forms, e.g., Forms 10-K, 10-Q, 8-K, 20-F, 6-K, and S-4 (see Table \ref{tab:form_types}).\footnote{Exhibit types are listed in 17 C.F.R. § 229.601.} Today, submission are typically made in the form of a .htm file, although older submissions were often made as .txt files (see Table \ref{tab:performance_table}).

We collected all 3,529,347 EDGAR filings from January 1, 2000, to March 21, 2023. Based on these filings, we then extracted those with exhibit numbers that indicate a possible material contract.

We focus on three types of exhibits: 2 (plans of acquisition or arrangement), 10 (material contracts), and 99 (additional exhibits). Although exhibit code 99 may not be the proper designation for material contracts, we found that it included many contracts. We include all files listed as exhibits 2 or 10, and for exhibit 99 files, we only retain those where the exhibit description indicates the file is an agreement, contract, merger plan, or other covenant. For the total 3,529,347 EDGAR filings, we identified 1,254,161 filed contracts. After adjusting for filings where the identified contract exhibits point to the same URL, we have 1,038,766 unique contract URLs. These 1,038,766 agreements represent partial de-duplication, since there is no further de-duplication based on the text of the agreement. Because filing meta-data may differ between filings and we use that meta-data in calculating our statistics, the statistics below are calculated based on the 1,254,161 filed agreements; with this approach, we do not have to decide which filing information to use when two filed contracts point to the same underlying URL. 
Tables \ref{tab:form_types} and \ref{tab:performance_table} describe the breakdown of the corpus based on filing form type and document extension of contract filing respectively. 
\begin{table}[h]
\centering
\begin{tabular}{lrr}
\hline
\textbf{Form Type} & \textbf{Count} & \textbf{Percent} \\
\hline
10-K    & 256,492  & 20.5\%\\
10-Q    & 265,556  & 21.2\%\\
20-F    & 8,336   & 0.7\% \\
40-F    & 502   &  0.0\% \\
6-K     & 10,729  & 0.9\% \\
8-K     & 503,722  & 40.2\% \\
DEF 14A & 8,401   & 0.7\% \\
S-4     & 200,423 & 16.0\% \\
\hline
\end{tabular}
\caption[Filings by form type]{Filings by form type. This table summarizes the number of filings for each specific SEC form type included in our dataset.}
\label{tab:form_types}
\end{table}

\begin{table}[ht]
\centering
\begin{tabular}{lrr}
\hline
\textbf{File Type} & \textbf{Count} & \textbf{Percent}\\
\hline
htm   & 1,026,698 & 81.9\%\\
html  & 386     & 0.0\%  \\
pdf   & 3,925    & 0.3\% \\
txt   & 223,151 & 17.8\%  \\
xfd   & 1      &0.0\%   \\
\hline
\end{tabular}
\caption[Filings by file type]{Filings by file type. This table summarizes the file type, given by document extension, across the contract exhibits included in our dataset.}
\label{tab:performance_table}
\end{table}

\subsection{Evaluating Agreement Types}

We hand-labeled 1,993 randomly-selected contracts with one of eight broad category labels.\footnote{Our labels further sub-divide contracts into one of 43 sub-types. These sub-types were grouped into the eight broader categories; training and labeling is based on these eight broader categories.} These eight categories and their contract frequencies are listed in Table \ref{tab:category_mapping}.
\begin{table}[ht!]
\centering
\begin{tabularx}{\textwidth}{@{}l c X@{}}
\toprule
\textbf{Agreement Type} & \textbf{No. Labels (\%)} & \textbf{Description} \\ \midrule
Security               & 529 (26.6\%) & Agreements related to a company's raising of debt and equity or transacting in derivatives, including credit agreements and guarantee agreements \\\\
Employment             & 794 (39.9\%) & Agreements between a company and key personnel, such as executives and board members \\\\
Lease                  & 63 (3.2\%) & Lease agreements \\\\
Services/Supply        & 130 (6.5\%) & Agreements with external firms or parties related to the provision of services or the sale of goods \\\\
Purchase/M\&A         & 227 (11.4\%) & Purchase agreements and M\&A agreements, including purchases of licenses, definitive merger agreements, and asset purchase agreements \\\\
Shareholder/Governance & 99 (5.0\%)  & Agreements related to formation of equity interests, governance of a company, and shareholder rights agreements \\\\
Other       & 88 (4.4\%)  & Other agreements not falling into one of the above categories,  \\\\
NA                     & 63 (3.2\%)  & Not a contract \\ \bottomrule
\end{tabularx}
\caption[Agreement Type and Labels]{Number and Description of Labels. This table describes the categories of agreement types captured in our dataset and the number of hand-labeled examples used to train the classification algorithm.}
\label{tab:category_mapping}

\end{table}

Using these labels, we fine-tuned a 7-billion-parameter LLaMA-2 model by adding a classification layer on top of the pre-trained large language model (LLM)~\citep{touvron2023llama2openfoundation}. The model accepts an input of 1,024 tokens and outputs a score for each of eight possible categories.\footnote{Tokenization is the process of converting words into tokens that the model understands. Typically, tokens are sub-word units.} The highest score is selected as the model’s prediction. We obtained the model from huggingface.co, a platform that provides access to open-source transformer models. For more efficient training, we adapted the model using Parameter-Efficient Fine-Tuning (PEFT), specifically employing Low-Rank Adaptation (LoRA)~\citep{hu2021loralowrankadaptationlarge}.

We assessed the performance of the fine-tuned model using five-fold cross-validation, withholding one-fifth of the labeled dataset as a validation set. Average performance across the folds was 95\% accuracy, with a weighted F1 score of 0.95. Performance across the individual folds was largely consistent. The performance of individual categories is reported in Table \ref{tab:avg_perf}, averaged over the five folds. The model’s strong overall accuracy demonstrates the efficacy of fine-tuning even relatively small LLMs. Performance across all categories was generally robust—achieving F1 scores of at least 0.89—with the exception of the “Other” category. We hypothesize that the “Other” category includes many contracts that are unique within the labeled training dataset, making it challenging for the model to achieve similarly strong performance.

\begin{table}[h!]
\centering
\begin{tabular}{l c c c}
\toprule
\textbf{Category} & \textbf{Precision} & \textbf{Recall} & \textbf{F1 Score} \\
\midrule
Security                & 0.95 & 0.95 & 0.95 \\
Employment              & 0.99 & 0.99 & 0.99 \\
Lease                   & 0.97 & 1.00 & 0.98 \\
Services/Supply         & 0.87 & 0.91 & 0.89 \\
Purchase/M\&A           & 0.88 & 0.89 & 0.89 \\
Shareholder/Governance  & 0.91 & 0.94 & 0.92 \\
Other                   & 0.85 & 0.74 & 0.79 \\
NA                      & 0.92 & 0.88 & 0.89 \\
\bottomrule
\end{tabular}
\caption[Average Performance Metrics by Category.]{Average Performance Metrics by Category. This table describes the precision, recall and F1 score for each of the categories. The tabulated values are average results from five-fold cross-validation.}
\label{tab:avg_perf}
\end{table}

After training the model, we deploy it across the entire dataset by assigning to each contract the category with the highest predictive score. These labels represent the agreement types reported in the Material Contract Corpus.

We then distinguish between different types of agreements—such as amendments, restatements, joinders, and terminations. This distinction is important because some categories may not include the full terms of the original contract. For example, an amendment may specify only the text to be modified, whereas an amendment that also serves as a restatement includes the entirety of the new agreement. Joinders and terminations generally do not restate the complete original contract. By labeling contracts in the MCC with these distinctions, users can more easily select contracts that are more likely to contain all the terms of the agreement. We use a simple keyword-matching algorithm applied to the initial contract text to determine whether a contract falls into one of these categories.\footnote{A fine-tuned LLM failed to achieve similar performance as the simple keyword matching algorithm.} Table \ref{tab:type-score} shows the performance of our algorithm on sets of randomly selected and hand-labeled contracts.

\begin{table}[htbp]
\centering
\resizebox{\textwidth}{!}{%
\begin{tabular}{lrrrr}
\hline
\textbf{Type} & \textbf{No. Labels} & \textbf{No. Pos. Labels} & \textbf{Pos. Label F1} & \textbf{Overall Accuracy} \\
\hline
Restatement  & 353 & 39  & 0.89 & 0.98 \\
Amendment    & 329 & 103 & 0.99 & 0.99 \\
Joinder      & 118 & 15  & 1.00 & 1.00 \\
Termination  & 117 & 7   & 1.00 & 1.00 \\
\hline
\end{tabular}%
}
\caption[Summary of labeled data and performance metrics.]{Summary of labeled data and performance metrics. We also label agreements across four binary categories that track whether an agreement is a restatement, amendment, joinder, or termination agreement. This categories are not mutually exclusive. This table reports the label support, F1 score, and accuracy of automated classifiers across these four binary categories.}
\label{tab:type-score}
\end{table}

\subsection{Documenting Contract Parties}

The MCC reports the parties associated with each contract, enabling researchers and practitioners to search for contracts linked to a given party or between two specified parties.

To identify the parties in a contract, we first deploy a pre-trained Named Entity Recognition model based on the RoBERTa architecture obtained from huggingface.co to extract named entities representing organizations or individuals~\citep{liu2019robertarobustlyoptimizedbert}. We tuned the algorithm to exclude named entities that are unlikely to be actual contract parties—for example, generic terms such as “board of directors,” “Delaware corporation,” and state names.\footnote{Matches are excluded when an entire tagged entity is within a set of generic entity names, such as those named in the text.} We tested the accuracy of our approach on a randomly selected sample of 100 contracts and achieved perfect recall, correctly identifying 100\% of all parties. In 96\% of the contracts, the results included only actual parties, while 4\% included one or more extraneous entities. Overall, these extra entities accounted for only 1.2\% of all tagged entities. In addition to the entities recognized in the contract text, we also added the filer name as a contract party when it was not already present.

The entities identified through this approach are not uniquely associated with a single legal entity; for example, Bank of America may appear in contracts as “Bank of America,” “Bank of America NA,” or “BoA.” To address the plurality of possible names for a given legal entity, we developed an approach to detect when two different names refer to the same entity and to link those names within the corpus. This approach allows a search for “JP Morgan Chase” to retrieve contracts where the tagged entity is “JPMC.”

To link names, we first created a list of suffixes to ignore when comparing party names. These substrings are typically words with relatively low information content, such as those indicating legal entity type (e.g., corporation, Inc., LLC). We also split party names containing a “doing business as” (D/B/A) modifier into two separate entities. After excluding these ignored terms, we compared the remaining portions of the two strings using fuzzy string matching—a method that assesses the degree of similarity between words.\footnote{Our method uses the Levenshtein ratio, calculated as the Levenshtein edit distance between the two strings divided by the length of the longer string.} We also considered perfect acronym matches provided the acronym was at least five letters long.\footnote{If two parties co-occur in the same contract, we also consider them a match for acronyms of three letters. For example, if “UPS” and “United Parcel Service” occur in the same contract, they would be considered a match for each other.} Comparing all tagged parties against each other produced a similarity score between every pair of identified parties. Pairs with a sufficiently high similarity score were considered to refer to the same legal entity. The similarity threshold was selected through manual inspection of randomly selected groups. Matches were treated as transitive; for example, if “JPMC” and “J P Morgan Chase” were determined to refer to the same party, and “JP Morgan Chase” and “J P Morgan Chase” were likewise matched, then “JPMC” would also be considered equivalent to “JP Morgan Chase.” This transitivity allowed us to create sets of different party names that all refer to the same entity, and a manual review of these groups further filtered out incorrect matches.

\section{Description of the Material Contracts Corpus}\label{sec:sec3}

In this section, we provide descriptive statistics of the MCC. Overall, the MCC comprises 1,254,161 filed contracts, which consolidate into 1,038,766 unique contract URLs. Table \ref{tab:corpus-breakdown} details the composition of the MCC by agreement type, industry, and state of incorporation, while Table \ref{tab:summary-stats} presents summary statistics for the numerical variables. Over half of the contracts fall into the security, employment, or Purchase/M\&A agreement categories. Similarly, over half of agreements are filed by a company incorporated in Delaware. The industry composition of the MCC is less skewed, with retail trade, finance, manufacturing, and construction being the four largest industries with a combined share of 63.48\%. 

\begin{table}[ht]
\centering
\begin{tabular}{l r r}
\hline
\textbf{Agreement Type} & \textbf{No. of Agreements} & \textbf{Percent of Total} \\
\hline
Security                & 362,609   & 28.9\% \\
Purchase/M\&A           & 141,667   & 11.3\% \\
Employment              & 486,406   & 38.8\% \\
Lease                   & 36,710    & 2.9\% \\
Services/Supply         & 89,840    & 7.2\% \\
Shareholder/Governance  & 55,881    & 4.5\% \\
Other                   & 46,149    & 3.7\% \\
N/A                      & 34,899    & 2.8\% \\
\hline\hline
\textbf{SIC Code} & \textbf{No. of Agreements} & \textbf{Percent of Total} \\
\hline
0 (Agr., Forestry, and Fishing) & 3,921    & 0.31\% \\
1 (Mining)                             & 79,929   & 6.38\% \\
2 (Construction)                       & 170,181  & 13.59\% \\
3 (Manufacturing)                      & 209,803  & 16.75\% \\
4 (Trans., Comm., Utilities)  & 122,801  & 9.81\% \\
5 (Wholesale Trade)                    & 108,266  & 8.64\% \\
6 (Retail Trade)                       & 250,614  & 20.01\% \\
7 (Finance, Insurance, Real Estate)   & 164,464  & 13.13\% \\
8 (Services)                           & 63,606   & 5.08\% \\
9 (Public Administration)             & 2,665    & 0.21\% \\
n (Not Listed)                        & 77,911   & 6.22\% \\
\hline\hline
\textbf{State of Incorporation} & \textbf{No. of Agreements} & \textbf{Percent of Total} \\
\hline
DE           & 633,795  & 50.5\% \\
NV           & 90,847   & 7.2\% \\
MD           & 52,830   & 4.2\% \\
CA           & 26,137   & 2.1\% \\
FL           & 24,606   & 2.0\% \\
NY           & 21,162   & 1.7\% \\
PA           & 18,696   & 1.5\% \\
OH           & 17,590   & 1.4\% \\
TX           & 17,473   & 1.4\% \\
Other States  & 174,345  & 13.9\% \\
Not Listed   & 176,680  & 14.1\% \\
\hline
\end{tabular}
\caption{MCC Composition: Agreement Types, SIC Industries, and State of Incorporation: This table shows the breakdown of the corpus by agreement type, industry (first digit of SIC code), and state. The state data includes the top ten occurring states, with all other states grouped into a single row for brevity. Not listed indicates that data was not available from the filing. }
\label{tab:corpus-breakdown}
\end{table}

\begin{table}[ht]
\centering
\begin{tabular}{lrrrrr}
\hline
 & \textbf{Mean} & \textbf{Median} & \textbf{STD} & \textbf{Min} & \textbf{Max} \\
\hline
Year            & 2010   & 2010   & 6.25   & 2000 & 2023 \\
Quarter         & 2.4    & 2      & 1.13   & 1    & 4    \\
Amendment       & 0.29   & 0      & 0.45   & 0    & 1    \\
Restatement     & 0.09   & 0      & 0.29   & 0    & 1    \\
Joinder         & 0.005  & 0      & 0.07   & 0    & 1    \\
Termination     & 0.005  & 0      & 0.07   & 0    & 1    \\
Number of Parties & 3.07 & 3      & 1.83   & 1    & 32   \\
\hline
\end{tabular}
\caption{Summary Statistics: This table presents summary statistics on the numerical categories of the MCC. Amendment, restatement, joinder and termination are binary labels, so the mean represents the fraction of data with that label.}
\label{tab:summary-stats}
\end{table}

\subsection{Trends over Time}

\begin{figure}
\centering\includegraphics[width=1\linewidth]
{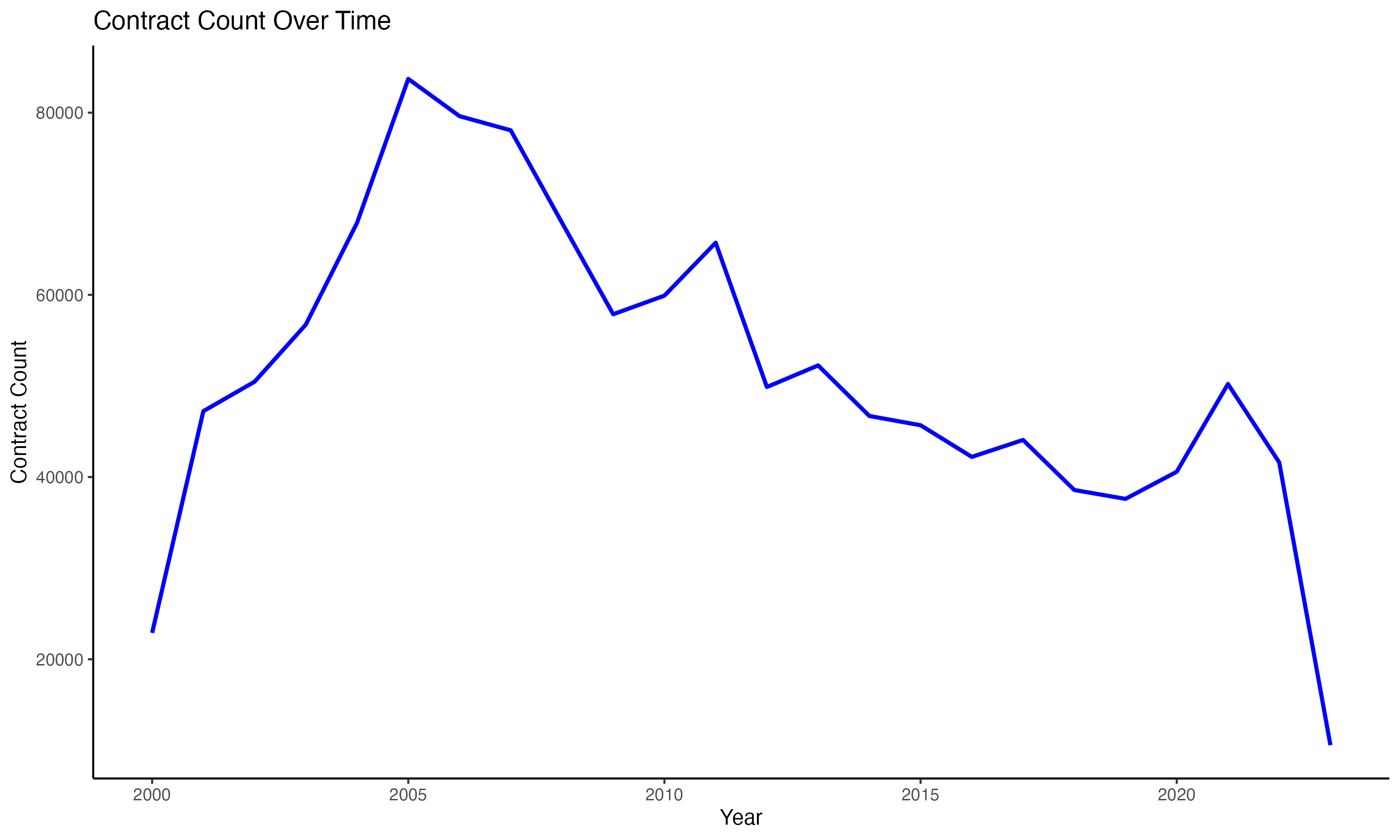}
\caption{Count of Contracts by Year: This figure shows the number of contracts filed on EDGAR. The drop in 2023 is due to corpus collection terminating in the first quarter of that year.}
\label{fig:overall_count}
\end{figure}

\begin{figure}[ht]
    \centering
\includegraphics[width=0.95\linewidth]{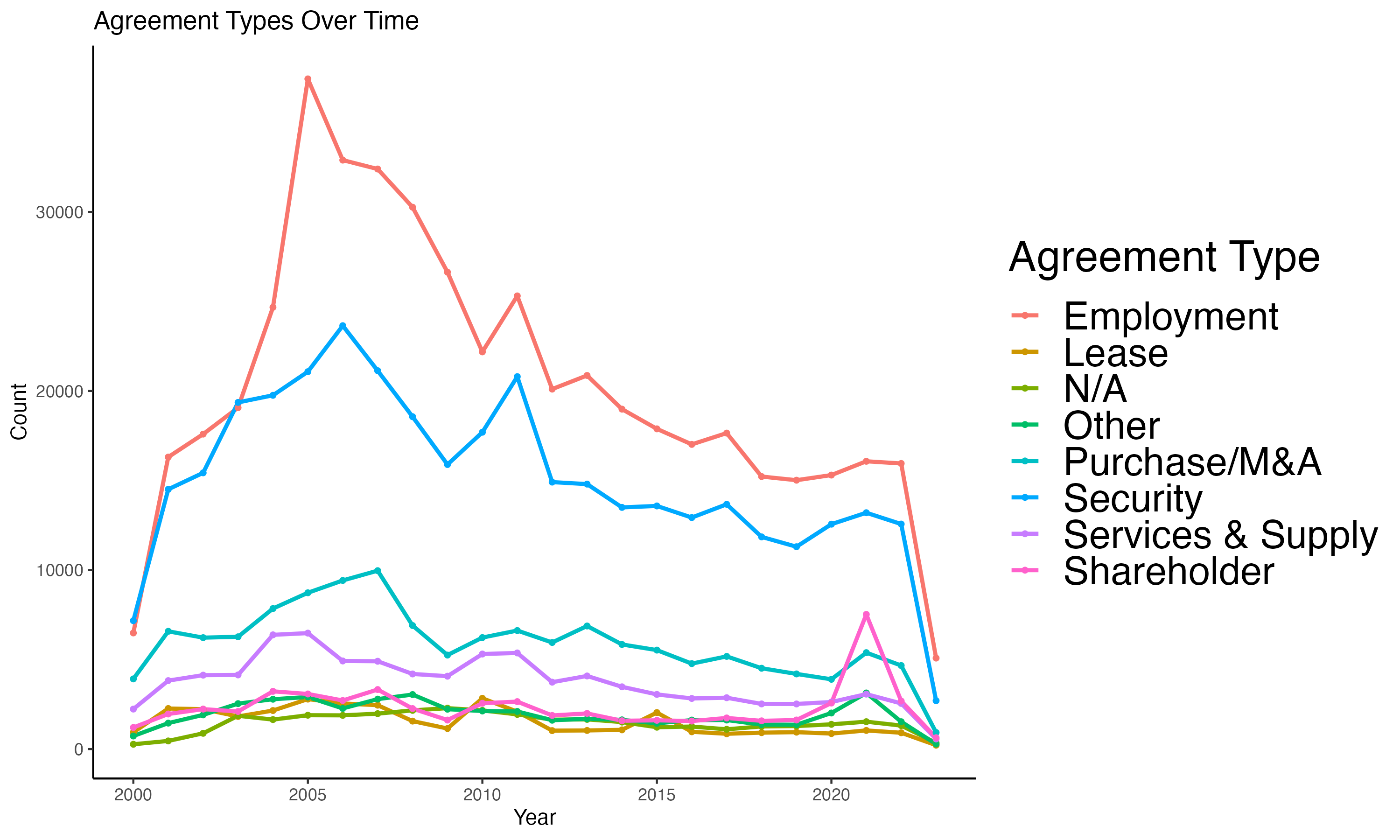}
    \caption{Count of Agreement Types by Year: This figure shows the number of agreements by agreement type by year. The dip in 2023 reflects the termination of the collection. The MCC's largest categories are employment and security agreements, reflecting the relative frequency of these agreements for public companies. The spike in shareholder agreements in 2021 coincides with a large increase in IPO activity.}
    \label{fig:agreement_types}
\end{figure}

Figure \ref{fig:overall_count} shows the total number of contracts filed each year. The drop in 2023 is due to the corpus collection ending in the first quarter of that year. Notably, material contract filings have declined since peaking in 2005. Figure \ref{fig:agreement_types} illustrates that agreement types are not equally represented across the corpus. In particular, \textit{employment}, which includes not only C-suite employment contracts, but also pension plans and other incentive mechanisms, are (virtually) consistently the most prevalent agreement type over time, followed by \textit{security}, which includes debt, credit, and guarantee agreements, among others. The \textit{shareholder} category saw a spike in 2021, likely driven by the sharp increase in Initial Public Offerings (IPOs) that year~\citep{ritter_ipo_update}.

% Combined subfigures for median word count and FK grade level
\begin{figure}[ht]
    \centering
    % First subfigure
    \begin{subfigure}[b]{0.48\linewidth}
        \centering
        \includegraphics[width=\linewidth]{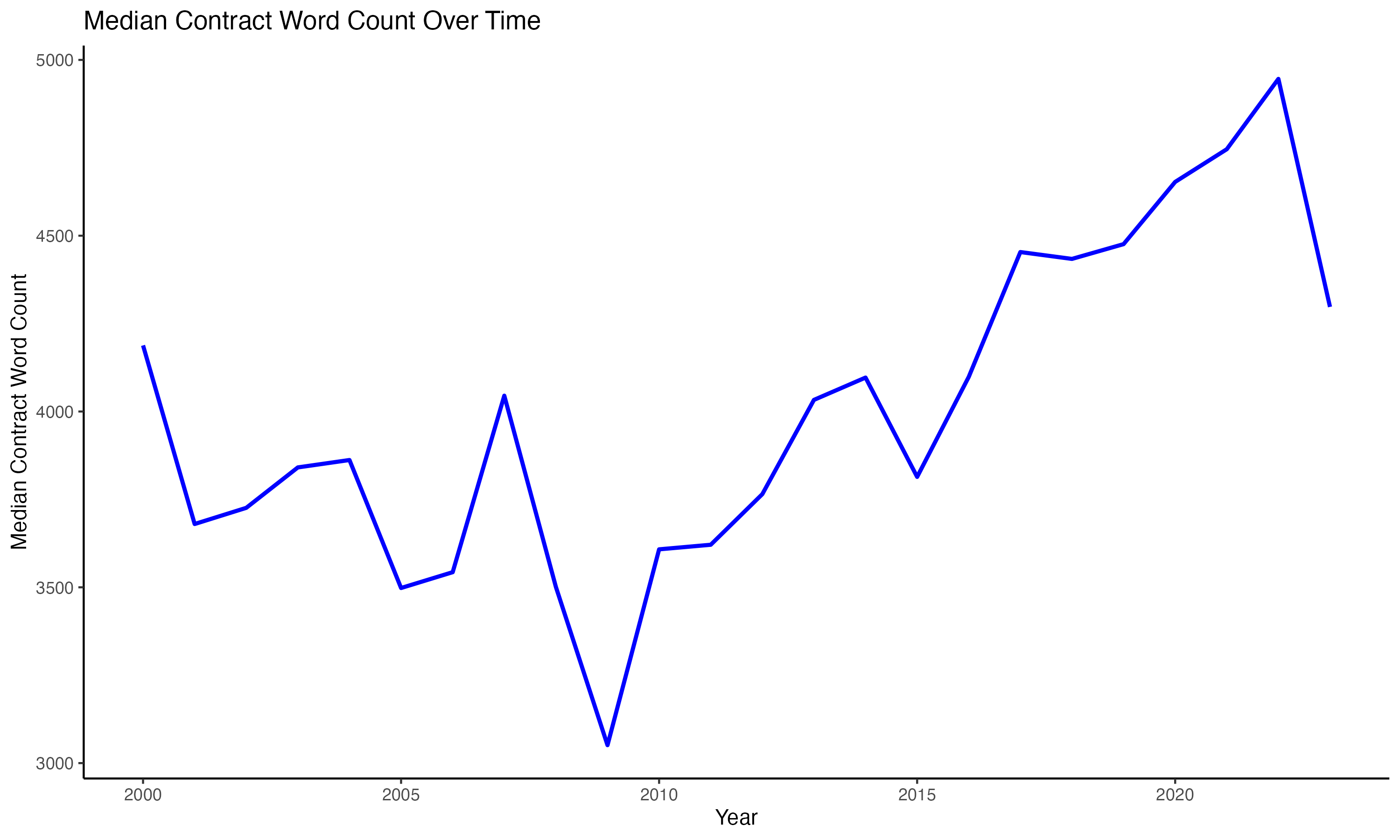}
        \caption{Median Word Count}
        \label{fig:overall_wc}
    \end{subfigure}
    \hfill
    % Second subfigure
    \begin{subfigure}[b]{0.48\linewidth}
        \centering
        \includegraphics[width=\linewidth]{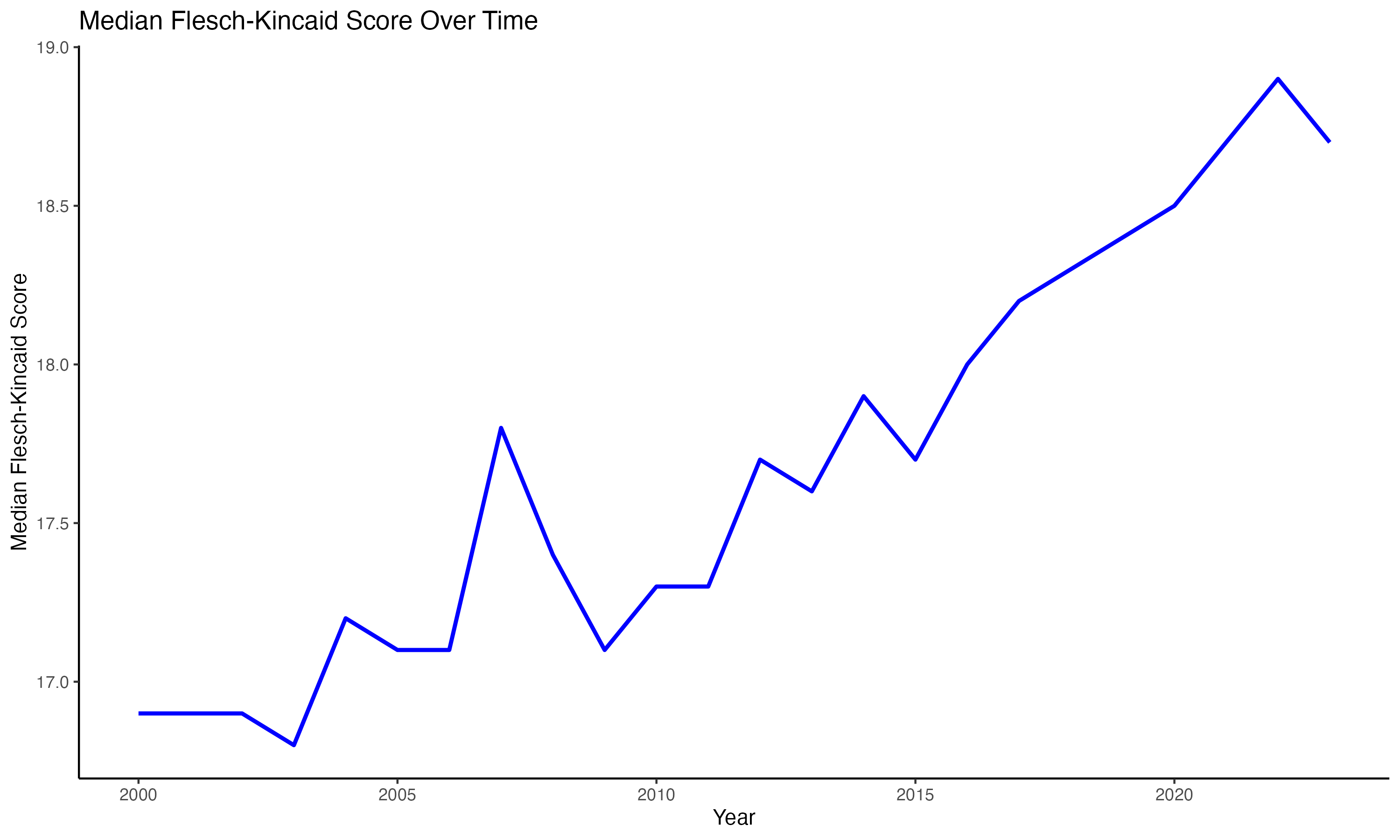}
        \caption{Median Flesch-Kincaid Grade Level}
        \label{fig:overall_fk}
    \end{subfigure}
    \caption{Median agreement length and readability by year: These figures show median agreement word count and Flesch-Kincaid grade level over time. Contracts have tended to increase in both length and reading complexity.}
    \label{fig:combined_metrics}
\end{figure}

Figure \ref{fig:overall_wc} shows that the length of publicly filed contracts has increased over time, and Figure \ref{fig:overall_fk} shows that the Flesch-Kincaid grade level of these contracts has also increased, suggesting that contracts are becoming more difficult to read. Detailed trends by agreement type are provided in Figures \ref{fig:wc_type}, \ref{fig:fk_type}, and \ref{fig:ttr_type} in the Appendix.  The observed trends are largely consistent with the findings of \citet{arbel2024readability}, who also discusses further limitations of readability measures and their interpretation.

\subsection{Parties}

\begin{figure}
\centering\includegraphics[width=1\linewidth]
{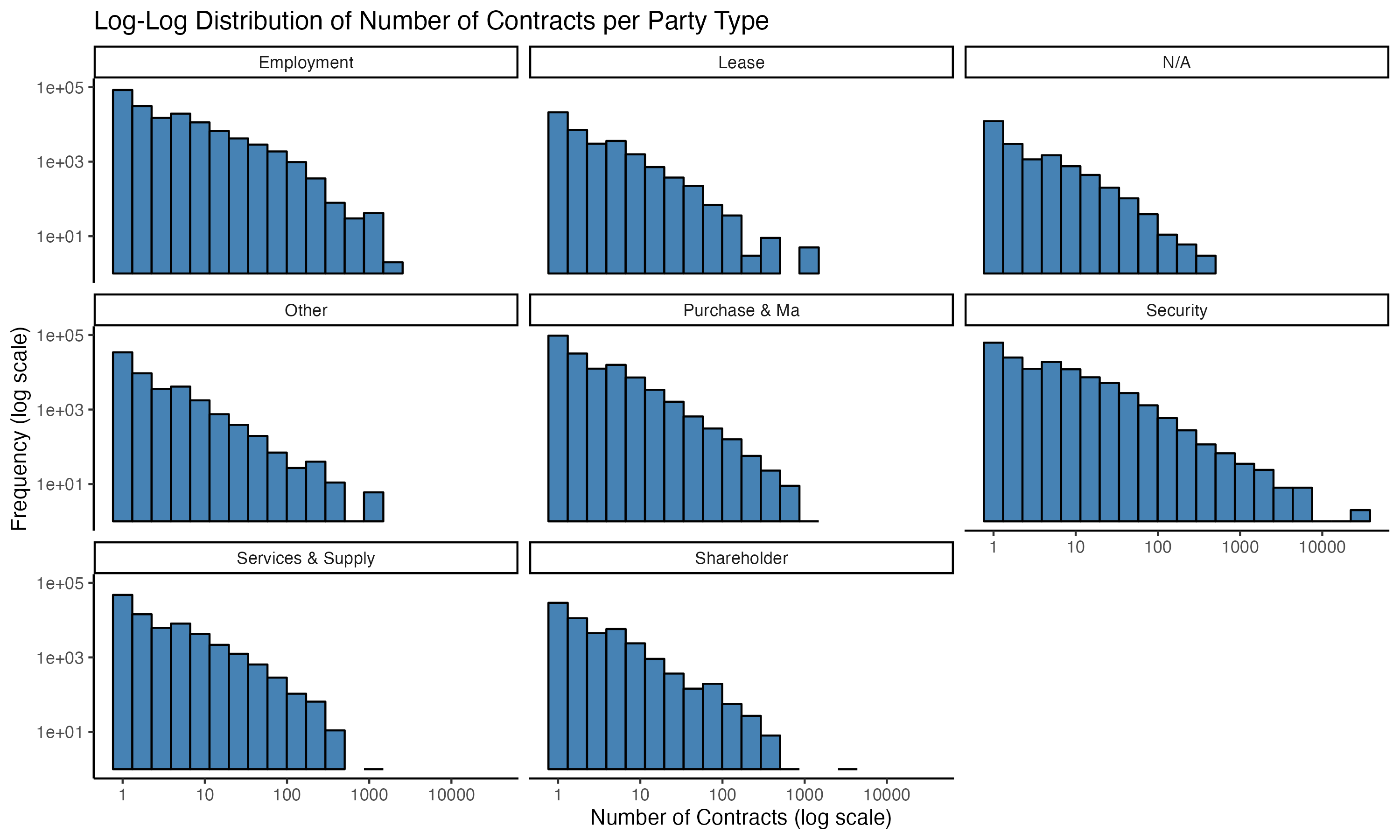}
\caption{Log-Log Histogram of Party Appearances by Agreement Type: This figure shows a histogram of party appearance by agreement type. The x axis represents the number of times an individual party appears in a contract for a given agreement type while the y axis shows the frequency of that appearance count. To adjust for skewness, the results are displayed on a log-log plot. All agreements types are highly skewed, as shown by the large number of parties featured in one contract.}
\label{fig:agreements_hist}
\end{figure}

The MCC encompasses 480,829 unique parties. The frequency of party appearances are highly skewed. Figure \ref{fig:agreements_hist} shows a histogram of party frequency; for example, about 10,000 parties appear only once for each type of agreement, while there are some parties that appear over 10,000 times in security agreements. The skewness of the data necessitates a log-log plot. Table \ref{tab:toptenparties} in the appendix presents the top ten parties along with their associated contract counts. Financial institutions are prevalent as contract parties, likely because the largest financial institutions frequently participate in financing agreements for public companies. The frequency distribution of parties is highly skewed: the 100th most frequent party is associated with 1,225 contracts, the 1,000th with only 273, and 233,064 parties appear only once in the MCC.

\section{Conclusion}

This paper presents the Material Contract Corpus (MCC), a comprehensive and publicly available dataset of over one million unique contracts filed with the SEC. By systematically extracting, classifying, and analyzing these contracts, we have demonstrated the feasibility and value of using advanced natural language processing techniques—such as fine-tuning a LLaMA-2 model to create important empirical datasets for use by scholars and practitioners. 

The MCC not only serves as a valuable resource for those interested in legal and contractual studies, but it also lays the groundwork for future work aimed at refining data extraction techniques, enhancing classification accuracy, and further exploring the dynamics of contractual language over time. 

\FloatBarrier
\printbibliography

\clearpage
\setcounter{page}{1}
\renewcommand{\thepage}{Appendix}
\appendix
\section*{Appendix}
\setcounter{figure}{0} \renewcommand{\thefigure}{A.\arabic{figure}}
\setcounter{table}{0} \renewcommand{\thetable}{A.\arabic{table}}

\begin{figure}[ht]
    \centering
    % First subplot
    \begin{subfigure}[b]{0.5\linewidth}
        \centering
        \includegraphics[width=\linewidth]{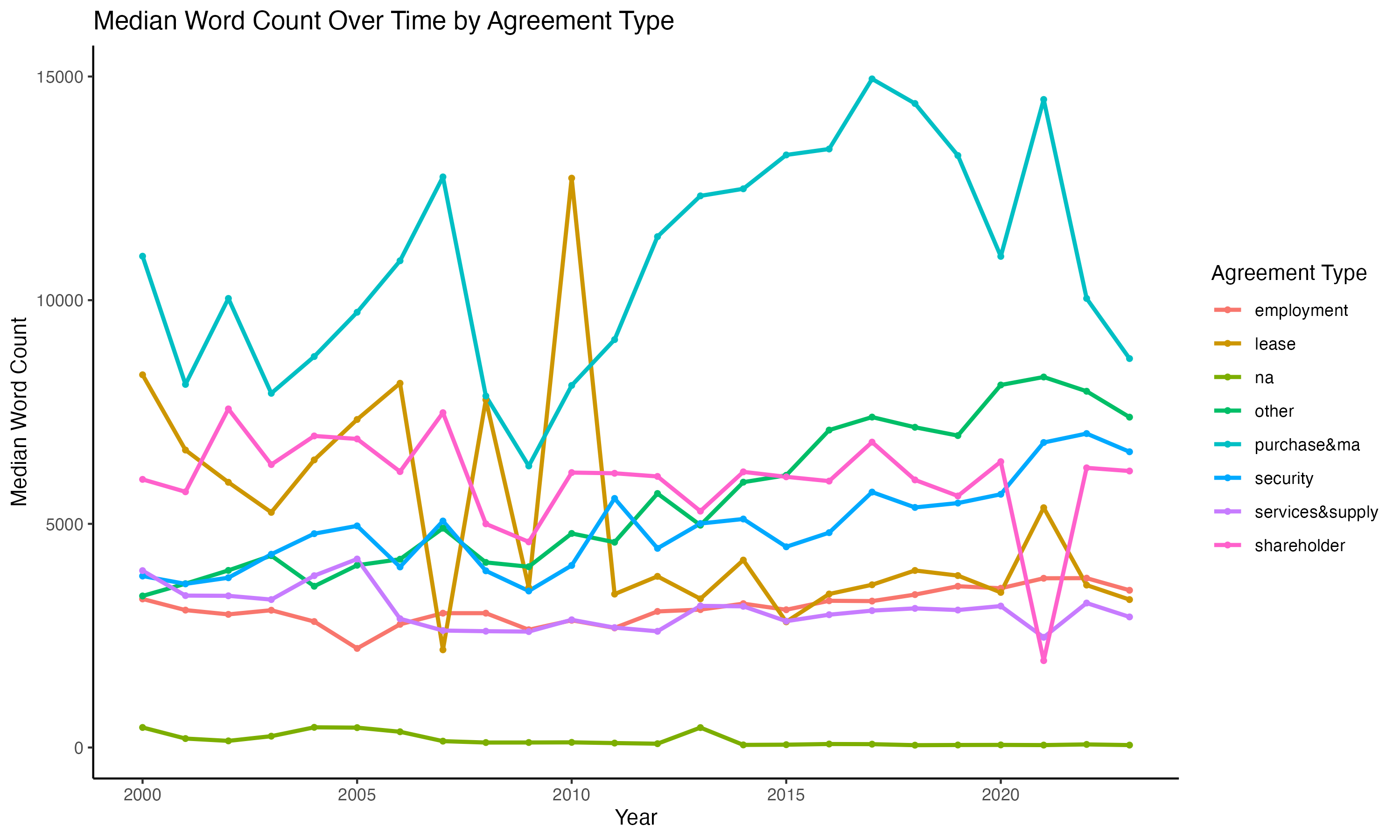}
        \caption{Median Agreement Word Count}
        \label{fig:wc_type}
    \end{subfigure}
    
    \vspace{1em} % vertical spacing between subplots
    
    % Second subplot
    \begin{subfigure}[b]{0.5\linewidth}
        \centering
        \includegraphics[width=\linewidth]{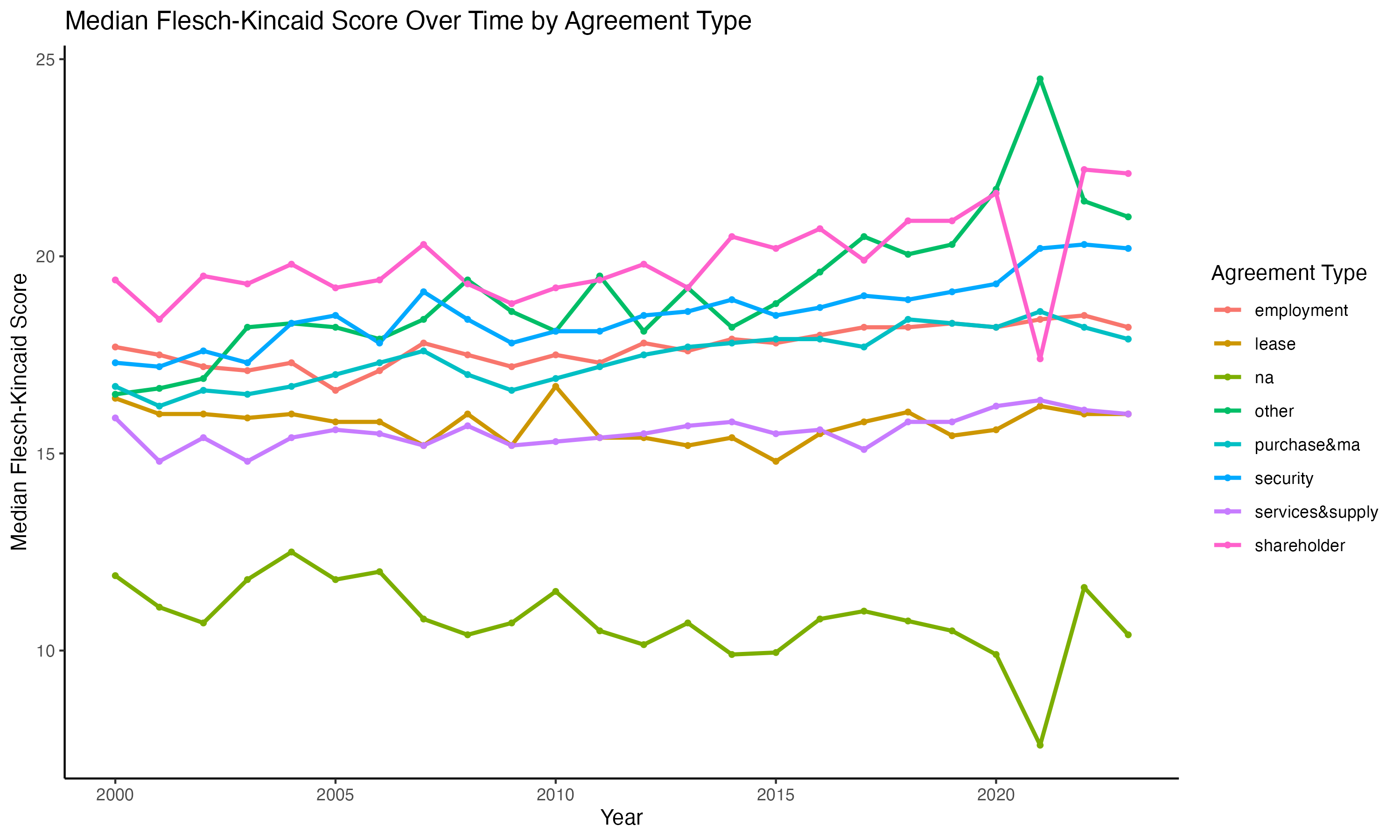}
        \caption{Median Contract Flesch-Kincaid Grade Level}
        \label{fig:fk_type}
    \end{subfigure}
    
    \vspace{1em} % vertical spacing between subplots
    
    % Third subplot
    \begin{subfigure}[b]{0.5\linewidth}
        \centering
        \includegraphics[width=\linewidth]{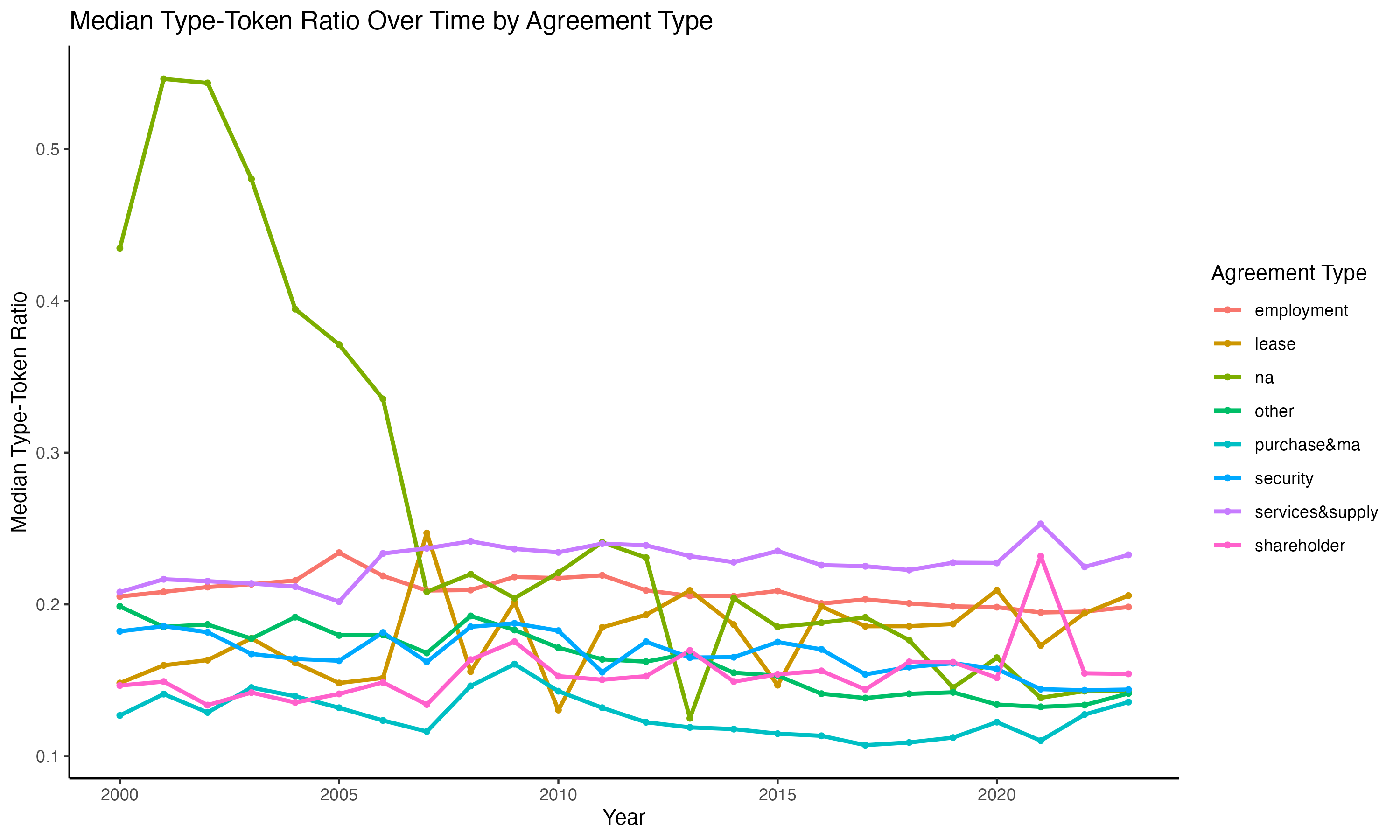}
        \caption{Median Contract TTR}
        \label{fig:ttr_type}
    \end{subfigure}
    
    \caption{Combined figure showing the median word count, Flesch-Kincaid grade level, and type-token ratio over time. M\&A agreements demonstrate the longest average length of agreement type while services and supply agreements are generally the shortest (excluding filings identified as non-contracts).}
    \label{fig:combined}
\end{figure}

\begin{table}[ht]
\centering
\begin{tabular}{l r}
\hline
\textbf{Party} & \textbf{Contracts Count} \\
\hline
Bank of America NA                  & 29,769 \\
JPMorgan Chase Bank NA              & 25,419 \\
Wells Fargo Bank National Association & 15,653 \\
US Bank National Association         & 10,235 \\
Deutsche Bank                        & 8,475  \\
JP Morgan Securities Inc             & 7,189  \\
Wachovia Bank National Association   & 6,594  \\
Citibank NA                          & 6,580  \\
Banc of America Securities LLC       & 5,748  \\
Select Medical Corporation           & 5,522  \\
\hline
\end{tabular}
\caption{Top Ten Parties in the Material Contracts Corpus: The prevalence of financial institution in this table reflects their frequent deals with other public companies.}
\label{tab:toptenparties}
\end{table}

\end{document}